\documentclass[letterpaper, 10 pt, conference]{ieeeconf} 

\usepackage{graphicx} 
\usepackage{algorithm}
\usepackage[utf8]{inputenc}
\usepackage{algpseudocode}
\usepackage{amsmath}

\usepackage[T1]{fontenc}
\let\labelindent\relax
\usepackage[inline]{enumitem}
\usepackage[table,xcdraw]{xcolor}
\usepackage{multirow}
\usepackage{color}
\usepackage{tabularray}
\usepackage{pgfplots}
\usepackage{tikz}
\usepackage{subcaption}
\pgfplotsset{compat=newest}
\usepackage{xcolor}
\usepackage{setspace}
\usepackage{amssymb}
\usepackage{mathtools}
\usepackage{adjustbox}
\usepackage[hidelinks]{hyperref}
\usepackage{booktabs}

\IEEEoverridecommandlockouts                             

\title{\LARGE \bf
Learning the Contact Manifold for Accurate Pose Estimation During Peg-in-Hole Insertion of Complex Geometries
}

\author{Abhay Negi$^{1}$, Omey M. Manyar$^{1}$, Dhanush Penmetsa$^{1}$, and Satyandra K. Gupta$^{1}$
\thanks{This work was supported by NASA Space Technology Graduate Research Opportunities Award (80NSSC24K1380).}% 
\thanks{$^{1}$Realization of Robotic Systems Lab, University of Southern California, Los Angeles, CA, USA. Address all correspondence to \href{mailto:guptask@usc.edu}{guptask@usc.edu}}}

\begin{document}

\maketitle
\thispagestyle{empty}
\pagestyle{empty}

%%%%%%%%%%%%%%%%%%%%%%%%%%%%%%%%%%%%%%%%%%%%%%%%%%%%%%%%%%%%%%%%%%%%%%%%%%%%%%%%
\begin{abstract}

Contact-rich assembly of complex, non‑convex parts with tight tolerances remains a formidable challenge. Purely model‑based methods struggle with discontinuous contact dynamics, while model‑free methods require vast data and often lack precision. In this work, we introduce a hybrid framework that uses only contact‑state information between a complex peg and its mating hole to recover the full $SE(3)$ pose during assembly. In under 10 seconds of online execution, a sequence of primitive probing motions constructs a local contact submanifold, which is then aligned to a precomputed offline contact manifold to yield sub‑mm and sub‑degree pose estimates. To eliminate costly k‑NN searches, we train a lightweight network that projects sparse contact observations onto the contact manifold and is 95× faster and 18\% more accurate. Our method, evaluated on three industrially relevant geometries with clearances of 0.1–1.0 mm, achieves a success rate of 93.3\%, a 4.1× improvement compared to primitive‑only strategies without state estimation. 

\end{abstract}

%%%%%%%%%%%%%%%%%%%%%%%%%%%%%%%%%%%%%%%%%%%%%%%%%%%%%%%%%%%%%%%%%%%%%%%%%%%%%%%%

\section{INTRODUCTION} 

\noindent Robotic assembly, particularly abstracted as the classic peg-in-hole problem, has been studied extensively for decades \cite{jiang_review_2022}. Despite significant advancements, the majority of industrial assembly tasks continue to rely heavily on human dexterity rather than robotic automation \cite{Ranjan_2024}. Current robotic assembly in industry remains predominantly restricted to structured environments and simple geometries, characterized by generous clearances (>1 mm) and minimal need for adaptive behavior. However, many critical industrial applications require mating complex, non-convex parts with extremely tight geometric tolerances \cite{mei_tase_2025, wang_contact_2018}. Under these conditions, traditional robotic methods often fail without accurate state estimation, leading to jamming, excessive force application, or even damage to components. To achieve broader adoption in complex industrial scenarios, current contact-rich manipulation methods must evolve to reliably handle challenging assemblies, ensuring efficiency, precision, and safety comparable to skilled human operators.

While the trade-offs between model-based and model-free approaches to robotic contact-rich assembly are well-understood and extensively explored \cite{jiang_review_2022}, addressing the assembly of complex, tight-tolerance geometries demands a more nuanced perspective - one that integrates the strengths of both learning and model-based approaches. Successfully performing complex, high-precision insertion tasks necessitates a strategically factored approach that leverages the structural insights of model-based methods while effectively employing learning to enhance generalization and computational efficiency.

Identifying appropriate modules where learning integration can yield significant advantages is critical. One approach is to utilize learning to estimate the parameters of contact models or insertion controllers, thereby improving adaptability \cite{tracy2025efficient, wu2023prim}. However, this is particularly challenging for non-convex geometries, where discontinuous contact dynamics can be prohibitive. In contrast, learning is more easily applied to accelerate computational bottlenecks within structured pipelines, such as projection, matching, or search, where data is easier to acquire. This integration improves scalability and real-time performance while preserving the transparency and reliability of model-based frameworks. In our work, we adopt this strategy by employing a learned metric projection function within a structured contact manifold framework.

Inspired by these principles and the human strategies in precision assembly tasks, we propose a novel hybrid framework that explicitly models a contact manifold, a structured representation of all feasible $\mathrm{SE(3)}$ poses at which the peg and hole can establish contact. Through offline, self-supervised exploration, we generate a comprehensive contact manifold that captures critical geometric and kinematic relationships during the making and breaking of contact. Online, our approach executes a concise sequence of primitive motions to construct a local sub-manifold of observed contact states. We then rapidly project this local sub-manifold onto the global manifold using our learned metric function, thereby achieving precise, interpretable state estimation without incurring the prohibitive computational costs of conventional kd-tree-based nearest neighbor queries.

Our method relies exclusively on proprioceptive kinematic sensing and force-based contact detection, avoiding reliance on additional sensing modalities. The explicit and interpretable nature of our state estimation enhances robot decision-making transparency and demonstrably reduces exerted forces during assembly, satisfying the stringent industrial requirements for efficiency, safety, and robustness. Our specific contributions include: (1) a structured contact-manifold-based methodology for peg-in-hole insertion of complex geometries using minimal sensing modalities; (2) a computationally efficient and accurate learning-based framework to replace expensive nearest neighbor searches; and (3) a robust algorithm for estimating precise $\mathrm{SE(3)}$ transformations between observed contact sub-manifolds and the global contact manifold.

\begin{figure*}[ht]
    \centering
    \includegraphics[width=1.0\linewidth]{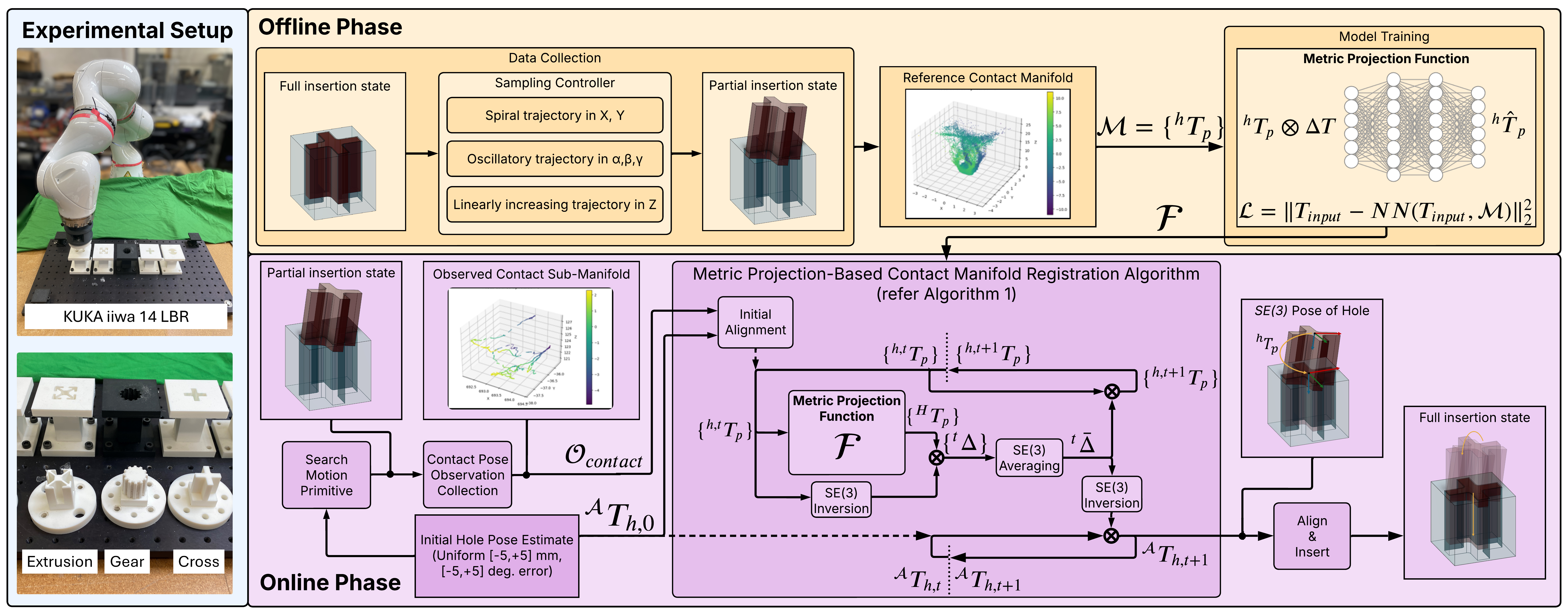}
    \caption{The overview of our methodology. During the offline phase, a sampling controller is used to sample contact poses and train a model, $\mathcal{F}$, to project neighboring points to the contact manifold. During the online phase, the system is provided an initial hole pose estimate with significant error. The search motion primitive uses this estimate to achieve a partial insertion state. Next, contact pose observation collection is performed by perturbing the peg pose, resulting in the observed contact submanifold, $\mathcal{O}_{contact}$. The contact manifold registration algorithm uses $\mathcal{O}_{contact}$, $\mathcal{F}$, and the initial hole pose estimate to compute a precise pose of the hole with respect to the robot. The system then uses this pose to align the peg and finally insert to achieve the full insertion state. $\mathcal{M}$ represents the contact manifold dataset, $NN$ represents nearest neighbor search. and $\boldsymbol{\otimes}$ represents composing SE(3) poses, corresponding to the multiplication of their homogeneous transformation matrices.}
    \label{fig:system-overview}   
\end{figure*}

\section{RELATED WORKS} 
\subsection{Model-Based Approaches} 
\label{ssec:rel-model-based}
\noindent Model-based methods \cite{lua_icra_2019} for peg-in-hole insertion leverage prior geometric knowledge to derive explicit contact models, reducing the task to learning or estimating model parameters. A common approach employs Dynamic Movement Primitives (DMPs) \cite{saveriano2023dynamic}, parameterized by harmonic functions, whose parameters are typically learned via human demonstrations \cite{wu2023prim} or reinforcement learning \cite{davchev_2022_ral}. Another common model-based technique is visual servoing \cite{haugaard_case_2022}, where learned visual features facilitate alignment. Significant work has focused on recognizing contact states \cite{jin_contact_2021, pankert_learning_2023-1} to guide assembly through structured transitions, enhancing robustness and adaptability. Our approach similarly exploits contact-state information; however, prior studies primarily address simple geometries with large clearances, leaving insertion of complex, tight-tolerance geometries largely unexplored.

\subsection{Model-Free Approaches}
\label{ssec:rel-model-free}
\noindent Model-free methods predominantly employ reinforcement learning (RL) to directly learn insertion policies without relying on explicit state information. These approaches typically require either high-fidelity simulators \cite{Narang-RSS-22}, \cite{tangAutoMateSpecialistGeneralist2024a}, or extensive offline demonstrations \cite{zhao_offline_2022, ball2023efficient} to train effective policies. Recent advancements in simulation techniques \cite{drake, zhu2023diff} have accelerated the popularity of RL-based methods; however, notable limitations remain \cite{jiang_review_2022}, including substantial data requirements \cite{singh2022reinforcement}, the complexity of sim-to-real transfer \cite{lee2024polyfit}, and reliance on multiple sensing modalities \cite{lee2020making}. These issues contribute to jamming and wedging, which remain prominent failure mode in practice \cite{tangIndustRealTransferringContactRich2023a}. Furthermore, despite their flexibility, these methods often suffer from slow execution, negatively impacting overall process efficiency. Moreover, few model-free approaches explicitly address the insertion of complex, tight-clearance geometries.

\section{METHODOLOGY} 

\noindent To solve for the pose of the hole with respect to the robot, we propose an approach illustrated in Fig. \ref{fig:system-overview}. Our methodology consists of two stages: an offline contact manifold metric projection learning phase and an online state estimation phase. In the offline phase, our goal is to learn a metric projection function, $\mathcal{F}$, which maps poses to the nearest pose on the contact manifold, $\mathcal{M}$. Specifically, $\mathcal{M}$ is a 6-D manifold comprising of the poses represented by $x,y,z$ position and Euler angles $\alpha,\beta,\gamma$, where the peg and hole are in contact. To construct $\mathcal{M}$, we fix the hole geometry at a known pose relative to the robot frame ($\mathcal{A}$) and systematically command the robot to explore poses where the peg makes contact with the hole. This process is detailed in Section \ref{ssec:reference-manifold-generation}. We then learn $\mathcal{F}$ by training a MLP to regress the nearest neighbor on $\mathcal{M}$ given an input pose. This process is detailed in \ref{ssec:metric-projection}. 

During the online phase, the objective is to estimate the relative transformation $^\mathcal{A}T_h$ by collecting contact observations $\mathcal{O}_{contact}$ and computing the corresponding contact submanifold. Our method builds on the iterative closest point (ICP) algorithm to iteratively estimate $^\mathcal{A}T_h$, aligning the observed contact submanifold with  the reference contact manifold $\mathcal{M}$ (see Section \ref{ssec:submanifold-generation}). The experimental setup consists of a KUKA LBR iiwa 14 R820 robot, which operates in one of two control modes ($\Omega$): (1) Joint Position Control or (2) Cartesian Impedance Control. The peg geometries (see Fig. \ref{fig:system-overview}) are rigidly attached to the robot's end-effector. The robot's integrated joint torque sensors provide end-effector wrench data, which is used to identify the contact poses (refer Section. \ref{ssec:submanifold-generation}). We assume an initial hole pose uncertainty sampled from a uniform distribution of [-5,+5] mm in translation (x,y) and [-5,+5] degrees in rotation ($\alpha$,$\beta$,$\gamma$). The online phase process flow is depicted in Fig. \ref{fig:system-overview}. 

\subsection{Reference Contact Manifold Generation}
\label{ssec:reference-manifold-generation}
\noindent Contact-based insertion strategies rely on a reference model of possible contact configurations. We define this as the reference contact manifold, $\mathcal{M}$, which captures the geometric relationship between the peg $p$ and hole $h$ during contact. The structure of $\mathcal{M}$ is intrinsically determined by the geometries of the interacting parts. Each distinct geometry results in a unique "fingerprint" in the 6D pose space encoding its contact characteristics. For example, square pegs generate manifolds with sharp transitions at corners and linear segments along edges, while cylindrical pegs produce radially symmetric, continuous manifolds. This correspondence between physical geometry and manifold structure is central to our method, enabling pose estimation via manifold registration.

\noindent\textbf{Sampling Methodology}
We represent the contact manifold as a set of discrete poses $^{h}T_{p}$ at which contact occurs. To sample this manifold, we employ a motion primitive that combines spiral trajectories in translation with oscillatory motions in rotation. These motions are executed with varying amplitude parameters depending on insertion depth in order to prevent part damage. Low-stiffness impedance control enables natural compliance during contact, and only poses where the measured contact force exceeds a threshold ($F_{ext} > \epsilon_f$) are recorded. To mitigate sampling bias, we incorporate random perturbations in the exploration trajectories. While this work utilizes physical interaction for data collection, the contact manifold can alternatively be constructed using simulation environments with accurate contact models \cite{kovalPoseEstimationPlanar2015b, kovalManifoldParticleFilter2017}.

\subsection{Learning-Based Metric Projection} \label{ssec:metric-projection} 

\noindent A metric projection is a function which maps each element of a metric space to the set of points nearest to that element in some fixed sub-space. In this work, we train a MLP to map a pose to the nearest pose on the contact manifold. The MLP input and output are 6-D poses represented as $T = \left[ x,y,z,\alpha,\beta,\gamma\right]^T \in \mathbb{R}^6$. The MLP is trained with input poses, $T_{input}$, sampled from $\mathcal{M}$ and offset by a bounded uniformly random offset, $T_{input} = ^hT^{\mathcal{M}}_p \otimes \Delta T$. The MLP is trained to output, $T_{output}$, minimizing the loss function, $\mathcal{L} = \| T_{input} - NN(T_{input}, \mathcal{M}) \|_2^2$, where $NN(T_{input}, \mathcal{M})$ is the nearest neighbor of $T_{input}$ from the set of poses in $\mathcal{M}$. Since we operate within offsets where the small-angle assumption is valid, we assume the manifold is locally Euclidean and utilize the Euclidean distance metric in the loss function. The MLP architecture comprises four hidden layers, each with a width of 4096 neurons, connected by ReLU activation functions, and is trained with the standard Adam optimizer. 

\subsection{Contact Observation Collection}
\label{ssec:submanifold-generation}
\noindent After learning $\mathcal{F}$ offline, we collect contact observations $\mathcal{O}_{contact} = \{^{\mathcal{A}}T_p\}$ at test time to estimate the unknown hole pose. All observations collected are of contact poses while the peg is engaged with the hole's inner walls - this partial insertion state is achieved with an oscillatory search motion primitive. Observation collection proceeds in two stages within a fixed time budget $t_{obs}$, combining spiral trajectories in the X and Y directions with oscillatory motions in the angular dimensions $\alpha$, $\beta$, and $\gamma$. The first stage uses large-amplitude motions to reach a target depth; the second uses smaller amplitudes to exploit the constrained configuration space at deeper insertions, yielding denser, more informative contact samples near the true alignment axis. 

Throughout both stages, the peg's pose $^{\mathcal{A}}T_{p}$ and wrench measurements are continuously recorded, with contact observations filtered using a force threshold: $\mathcal{O}_{contact} = \{ ^{\mathcal{A}}T_{p} \mid F_{ext} > F_{threshold}\}$. These filtered observations form a contact submanifold in $\mathcal{A}$'s frame, which is subsequently registered against the reference manifold $\mathcal{M}$ to estimate $^{\mathcal{A}}T_h$. Given the temporal continuity of these observations, we adopt a temporal downsampling procedure to a fixed number of points, which simultaneously increases variance in the sample set and reduces computational cost for registration. 

\subsection{Pose Estimation} 
\label{ssec:pose-estimation}

\begin{algorithm*}[ht] 
\setstretch{1.15}
\caption{Metric Projection-Based Contact Manifold Registration Algorithm} 
\label{alg:pose_icp}
\begin{algorithmic}[1]

    \State \textbf{Input:} $\{^\mathcal{A}T_p\}$, $^\mathcal{A}T_{h,0}$ \algorithmiccomment{Observed contact submanifold \& initial estimate} 
    
    \State $\{^{h,0}T_{p}\} \gets \left( ^{\mathcal{A}}T_{h,0} \right)^{-1} \otimes \{ ^{\mathcal{A}}T_{p} \}$ \algorithmiccomment{Initial alignment} \label{line:initial alignment}
    
    \For {$t = 0$ \textbf{to} $N-1$} 

        \For {$^{h,t}T_{p,i} \in \{^{h,t}T_p\}$} 

            \State $^{H}T_{p,i} \gets \boldsymbol{\mathcal{F}} \left(^{h,t}T_{p,i}\right)$  \algorithmiccomment{Compute projection onto reference manifold} 

            \State $^t\Delta_i \gets \left( ^{h,t}T_{p,i}\right)^{-1} \otimes \left( ^HT_{p,i} \right) $ \algorithmiccomment{Compute misalignments}     
            
        \EndFor

        \State $ ^{h,t}T_{h,t+1} \gets SE(3) \text{ mean pose of } \{ ^t\Delta \} $ \algorithmiccomment{Compute aggregate pose update from misalignments} 

        \State $  ^\mathcal{A}T_{h,t+1}  \gets \left( ^\mathcal{A}T_{h,t} \right) \otimes \left( ^{h,t}T_{h,t+1} \right)$   \algorithmiccomment{Update hole pose estimate} 
        
        \State $\{ ^{h,t+1}T_{p} \} \gets \left( ^{h,t}T_{h,t+1} \right)^{-1} \otimes \{ ^{h,t}T_{p} \} $ \algorithmiccomment{Update observations}
                    
    \EndFor 

    \State \textbf{return} $^{\mathcal{A}}T_{h,N}$
    
    \end{algorithmic}
\end{algorithm*}
\setstretch{1}

\noindent{The $SE(3)$ pose of hole $h$ with respect to the agent $\mathcal{A}$ is estimated by utilizing the metric projection-based contact manifold registration algorithm outlined in Algorithm \ref{alg:pose_icp}. The algorithm iteratively determines the $SE(3)$ pose that aligns the observed contact submanifold, $\mathcal{O}_{contact}$, with the reference contact manifold, $\mathcal{M}$. The contact observations and correspondences are each a point cloud, where each point is a contact pose represented using a position vector and Euler angles, i.e. $T = \left[ x,y,z,\alpha,\beta,\gamma\right]^T \in \mathbb{R}^6$. Similar to ICP in $\mathbb{R}^3$, the observations are iteratively aligned with the manifold by determining correspondences, computing misalignments, aligning the observations to the manifold, and updating the pose estimate. Alternatively, particle filters may be used for contact manifold-based pose estimation, though they require careful engineering to mitigate issues such as particle starvation and weight collapse \cite{kovalPoseEstimationPlanar2015b}, \cite{kovalManifoldParticleFilter2017}.}

The observed contact submanifold is a set of poses of the peg with respect to the agent, $\{ ^\mathcal{A}T_p \}$. The initial estimate, $^\mathcal{A}T_{h,0}$, is used to perform a coarse initial alignment of the set of observations to the hole frame, as shown in Line \ref{line:initial alignment}. Within each loop iteration, correspondence points on the contact manifold are found for each observation using the metric projection function, $\mathcal{F}$. The reference-to-observation misalignment, $\Delta_i$, is computed as the $SE(3)$ relative pose between the corresponding observation and reference point. 

After computing misalignments at each observation point, the aggregate misalignment—computed as the $SE(3)$ mean pose—is used to update the pose estimate and observations, and this process is repeated iteratively until a maximum number of iterations is reached. Using the final pose estimate, the peg is aligned and inserted by moving along its z-axis. 

\section{RESULTS} 

\begin{table}[ht]
\centering
\caption{Pose estimation and insertion success rate results}
\label{tab:results}
\begin{adjustbox}{max width=\linewidth}
\begin{tabular}{@{}lrrr@{}}
\toprule
\textbf{Geometry}                        & \textbf{Cross} & \textbf{Gear} & \textbf{Extrusion} \\ \midrule
Clearance [mm]                               & 0.1            & 0.3           & 1.0                \\
Number of Trials                         & 40             & 60            & 50                 \\
Translation Mean Abs. Err. {[}mm{]}      & 0.50           & 0.09          & 0.08               \\
Translation Err. Std. Dev {[}mm{]}       & 0.18           & 0.05          & 0.06               \\
Rotation Mean Abs. Err. {[}deg{]}        & 0.94           & 0.36          & 0.86               \\
Rotation Err. Std. Dev {[}deg{]}         & 1.17           & 0.22          & 0.91               \\
Success Rate - Direct Insertion {[}\%{]} & 22.5           & 20.0          & 26.0               \\
Success Rate - Our Method {[}\%{]}       & 80.0          & 100.0         & 100.0              \\ \bottomrule
\end{tabular}
\end{adjustbox}
\end{table}

\noindent\textbf{MLP-based Nearest-Neighbor Search: }We evaluate the regression performance of our MLP-based projection against a traditional kd-tree nearest neighbor search. Our learned metric function retrieves neighbors with MAE of 0.0454 mm and 0.0264$^\circ$ and error standard deviations of 0.0525 mm and 0.0296$^\circ$, in translation and rotation, respectively, between the MLP-predicted neighbor and the true nearest neighbor from the training dataset. Additionally, we compare the pose estimation speed and accuracy over 25 trials for 10 sets of observations and observe a reduction in mean computation time from 2.75 seconds to 29.0 milliseconds (100 observations, 50 iterations), and a mean reduction in MAE from 0.20 mm and 0.44$^\circ$ to 0.19 mm and 0.34$^\circ$ (excluding z). This result verifies the fidelity of the MLP used to approximate the metric projection function of input pose to the contact manifold. 

\noindent\textbf{Pose Estimation}
We conducted at least 40 experimental runs for each of three object geometries. As shown in Table \ref{tab:results}, pose accuracy was evaluated in both translation and rotation. Averaged across geometries, the \textbf{mean absolute position errors (X, Y)} was 0.22 mm, and the \textbf{orientation errors ($\boldsymbol{\alpha}$, $\boldsymbol{\beta}$, $\boldsymbol{\gamma}$)} was 0.72$^\circ$. Accuracy is higher in translation than in rotation due to the manifold having lower variance in the translational dimensions - this corresponds to the peg having less play in translation than in rotation. Notably, the cross geometry has significantly lower pose estimation accuracy - we hypothesize this to be due to insufficient coverage of the contact manifold during offline data collection. 

\noindent\textbf{Insertion Success Trials: }To evaluate insertion performance, we benchmarked our approach against a baseline in which the robot begins from a partial insertion state, executes a 10-second motion primitive to advance the insertion, and then performs a straight-line motion along the peg’s z-axis using impedance control (refer Table \ref{tab:results}). For fairness, impedance gains were tuned to maximize success, and each geometry was tested over 40 trials. Despite these efforts, the baseline approach yielded poor results, achieving an average success rate of only 22.8\% across geometries. These outcomes underscore the challenge of tight-clearance insertion for non-trivial shapes, where uncorrected misalignment often leads to jamming. In contrast, our method—leveraging contact pose estimation—consistently outperformed the baseline, achieving success rates of 80.0\%, 100.0\%, and 100.0\% for the respective geometries. Notably, the cross geometry exhibited the lowest success rate, likely due to reduced pose estimation accuracy. In contrast, the gear and extrusion geometries achieved 100\% success rates across a large number of trials, consistent with their higher pose estimation accuracy.

\section{CONCLUSION} 
\noindent We proposed a contact manifold-based approach for accurate pose estimation relying solely on pose and binary contact information. We demonstrated how learning can be leveraged in this structured approach by modeling a metric projection function that projects observations onto the contact manifold. We demonstrate an algorithm which iteratively aligns contact observations with the contact manifold. By registering the online sub-manifold to a precomputed global manifold, we achieved sub-mm/deg accuracy, improving success rates from 22.8\% to 93.3\%. Furthermore, by learning the metric projection function, we reduce computation time by 95× faster while reducing error by 18\%. Our approach is data-efficient, interpretable, and robust for tight-tolerance assembly. In future work, we plan to integrate learning methods for performing informed sampling of observations, and we plan to leverage simulation for generating the contact manifold. 

%%%%%%%%%%%%%%%%%%%%%%%%%%%%%%%%%%%%%%%%%%%%%%%%%%%%%%%%%%%%%%%%%%%%%%%%%%%%%%%%

\newpage 

%%%%%%%%%%%%%%%%%%%%%%%%%%%%%%%%%%%%%%%%%%%%%%%%%%%%%%%%%%%%%%%%%%%%%%%%%%%%%%%%

\bibliographystyle{ieeetr}
\bibliography{references}

\end{document}